\documentclass[sigconf]{acmart}
\usepackage{multirow,multicol}
\usepackage{bbm}

\AtBeginDocument{%
  }

\copyrightyear{2023}
\acmYear{2023}
\setcopyright{rightsretained}
\acmConference[MM '23]{Proceedings of the 31st ACM International Conference on Multimedia}{October 29-November 3, 2023}{Ottawa, ON, Canada}
\acmBooktitle{Proceedings of the 31st ACM International Conference on Multimedia (MM '23), October 29-November 3, 2023, Ottawa, ON, Canada}
\acmDOI{10.1145/3581783.3612703}
\acmISBN{979-8-4007-0108-5/23/10}

\makeatletter
\gdef\@copyrightpermission{
 \begin{minipage}{0.3\columnwidth}
  \href{https://creativecommons.org/licenses/by/4.0/}{\includegraphics[width=0.90\textwidth]{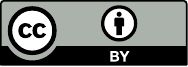}}
 \end{minipage}\hfill
 \begin{minipage}{0.7\columnwidth}
  \href{https://creativecommons.org/licenses/by/4.0/}{This work is licensed under a Creative Commons Attribution International 4.0 License.}
 \end{minipage}
 \vspace{5pt}
}
\makeatother

\acmSubmissionID{3}

\begin{document}

\title{Semantics2Hands: Transferring Hand Motion Semantics between Avatars}

\author{Zijie Ye}
\email{yzjscwy@gmail.com}
\affiliation{%
  \institution{Department of Computer Science and Technology, Tsinghua University}
  \city{Beijing 100084}
  \country{China}
}

\author{Jia Jia}
\email{jjia@tsinghua.edu.cn}
\authornote{Corresponding author.}
\affiliation{%
  \institution{Department of Computer Science and Technology, Tsinghua University}
  \institution{Beijing National Research Center for Information Science and Technology}
  \city{Beijing 100084}
  \country{China}
}

\author{Junliang Xing}
\email{jlxing@tsinghua.edu.cn}
\affiliation{%
  \institution{Department of Computer Science and Technology, Tsinghua University}
  \city{Beijing 100084}
  \country{China}
}

\renewcommand{\shortauthors}{Zijie Ye, Jia Jia, \& Junliang Xing}

\begin{abstract}
  Human hands, the primary means of non-verbal communication, convey intricate semantics in various scenarios. Due to the high sensitivity of individuals to hand motions, even minor errors in hand motions can significantly impact the user experience. Real applications often involve multiple avatars with varying hand shapes, highlighting the importance of maintaining the intricate semantics of hand motions across the avatars. Therefore, this paper aims to transfer the hand motion semantics between diverse avatars based on their respective hand models. To address this problem, we introduce a novel anatomy-based semantic matrix (ASM) that encodes the semantics of hand motions. The ASM quantifies the positions of the palm and other joints relative to the local frame of the corresponding joint, enabling precise retargeting of hand motions. Subsequently, we obtain a mapping function from the source ASM to the target hand joint rotations by employing an anatomy-based semantics reconstruction network (ASRN). We train the ASRN using a semi-supervised learning strategy on the Mixamo and InterHand2.6M datasets. We evaluate our method in intra-domain and cross-domain hand motion retargeting tasks. The qualitative and quantitative results demonstrate the significant superiority of our ASRN over the state-of-the-arts. Project page: \href{https://abcyzj.github.io/S2H/}{\textit{Semantics2Hands}}.
\end{abstract}

\ccsdesc[500]{Computing methodologies~Motion processing}
\ccsdesc[500]{Applied computing~Media arts}

\keywords{Hand Motion Retargeting, Neural Motion Processing}

\maketitle

\section{Introduction}

\begin{figure}[htbp]
  \centering
  \includegraphics[width=0.85\linewidth]{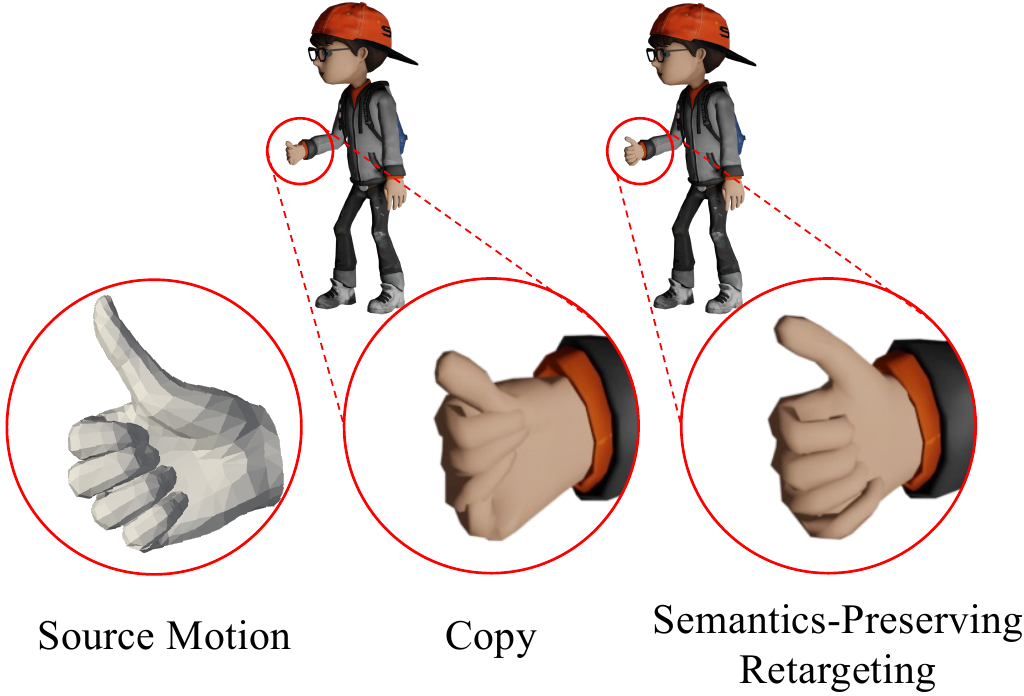}
  \caption{Despite the accurate body motions, errors introduced by copying finger joint rotations make the ``thumb-up'' gesture illegible.}
  \label{fig:motivation}
\end{figure}

The generation of realistic hand motions has demonstrated promising potential in diverse virtual avatar scenarios, including co-speech gesture synthesis~\cite{qi2023diverse,ye2023salient,ng2021body2hands} and sign language synthesis~\cite{zelinka2020neural,huang2021towards,saunders2022signing}. Human hands, being the primary means of non-verbal communication~\cite{studdert1994hand}, convey subtle nuances during the execution of particular hand gestures. Given people's high sensitivity to hand motions, even slight errors can significantly impact the user experience in virtual avatar applications. Consequently, maintaining consistent hand motion semantics across various virtual avatar hands is paramount. However, due to the highly articulated nature of the human hand with multiple degrees of freedom (DoFs) and the varying hand shapes and proportions of different avatars, directly copying joint rotations would significantly compromise the intricate semantics of hand motions, as shown in Figure~\ref{fig:motivation}. Consequently, developing a methodology that can preserve the semantics of hand motions when retargeting them to diverse avatars is essential.

Previous research has focused on motion retargeting and hand-object interaction. Motion retargeting, pioneered by \citet{gleicher1998retargetting}, aims to identify the characteristics of source motions and transfer them to target motions on different characters. Early work~\cite{lee1999hierarchical,bernardin2017normalized,feng2012automating} focused on optimization-based approaches. Recently, researchers have proposed data-driven approaches~\cite{villegas2018neural,zhang2023skinned,DBLP:journals/tog/AbermanLLSCC20} using various network architectures and semantic measurements. These approaches can successfully retarget realistic body motions but do not apply to dexterous hand motion retargeting. \citet{ge2005motion} proposed a rule-based approach for retargeting sign language motions; however, their method is limited to a specific set of pre-defined hand movements and lacks sufficient testing. Hand-object interaction is a research area that focuses on synthesizing realistic hand motions during interactions with objects, including static grasp synthesis~\cite{hasson2019learning,taylor2016efficient,zhu2021toward} and manipulation motion synthesis~\cite{DBLP:conf/eccv/ZhouBLP22,DBLP:journals/tog/ZhangYSK21,mordatch2012contact,ye2012synthesis}. However, these methods fail to preserve the semantics of hand motions in communication scenarios. Furthermore, they do not apply to diverse hand models with varying shapes and proportions. Despite the existing methods, it remains a challenge: retarget realistic hand motions with high fidelity across different hand models while preserving intricate motion semantics.

This paper focuses on retargeting dexterous hand motions across different hand models while preserving the semantics of the source hand motions. Hand motion retargeting requires a higher level of semantic measurement precision than body motion retargeting, making this idea novel. The semantic measurements previously employed in motion retargeting, including cycle consistency~\cite{villegas2018neural,DBLP:journals/tog/AbermanLLSCC20} and distance matrix~\cite{zhang2023skinned}, are inadequate due to the high density of hand joints within a limited space, which results in significant spatial interactions between finger joints and the palm.

Therefore, our central insight is that the spatial relationships between the finger joints and the palm are crucial for preserving hand motion semantics. Consequently, we encode the spatial relationships into a novel anatomy-based semantic matrix (ASM). We utilize ASM as the semantic measurement for precise hand motion retargeting. In particular, we first build anatomical local coordinate frames for finger joints on different hand models. Then we construct ASM based on the anatomical local coordinate frames. ASM quantifies the positions of the palm and other joints relative to the local frame of the given finger joint. Next, we acquire a mapping function from the source motion ASM to the target motion rotations using an anatomy-based semantics reconstruction network (ASRN). We train ASRN on two heterogeneous hand motion datasets~\cite{mixamo,Moon_2020_ECCV_InterHand2.6M}. Unlike template mesh-based methods~\cite{DBLP:conf/eccv/ZhouBLP22,DBLP:journals/tog/ZhangYSK21} for semantic correspondence, our approach is not dependent on template meshes and can be applied to various hand models.

We conducted comprehensive experiments to assess the quality of the hand motions generated by our ASRN. These experiments encompassed both intra-domain and cross-domain hand motion retargeting scenarios involving intricate hand motion sequences and a diverse range of hand shapes. The qualitative and quantitative results show that our ASRN outperforms existing motion retargeting methods by a large margin.

To summarize, our contributions are as three-fold:
\begin{itemize}
  \item We propose a novel task: semantics-preserving retargeting of dexterous hand motions across diverse hand models.
  \item We introduce an anatomy-based semantic matrix (ASM) that quantifies hand motion semantics without relying on any template mesh, making it applicable to various hand models.
  \item We propose a novel framework for semantics-preserving hand motion retargeting, leveraging the ASM. Experimental results on both intra-domain and cross-domain hand motion retargeting tasks validate the superior performance of our framework over existing methods.
\end{itemize}

\section{Related Work}

\subsection{Motion Retargeting}
Motion retargeting aims to identify the features of the source motions and transfer them to the target motions on a different character. The pioneering work by \citet{gleicher1998retargetting} addresses motion retargeting as a spatial-temporal optimization problem with the source motion features as kinematic constraints. Subsequent studies propose solutions to this optimization problem with various constraints~\cite{tak2005physically,lee1999hierarchical,choi2000online,bernardin2017normalized}.

Recently, data-driven methods~\cite{jang2018variational,delhaisse2017transfer,villegas2018neural,DBLP:journals/tog/AbermanLLSCC20,zhu2017unpaired,zhang2023skinned,lim2019pmnet} have become increasingly appealing due to the growing availability of motion capture data. \citet{delhaisse2017transfer} and \citet{jang2018variational} train neural networks for retargeting using paired training data. Subsequently, \citet{villegas2018neural} develop an adversarial neural network trained with cycle consistency~\cite{zhu2017unpaired}, eliminating the need for paired ground truth. \citet{DBLP:journals/tog/AbermanLLSCC20} propose a skeleton-aware network for retargeting motions between skeletons with varying topologies. \citet{zhang2023skinned} also introduces the Distance Matrix for measuring body motion semantics.

However, all the methods above either truncate finger movements or merely replicate finger joint rotations during retargeting, resulting in the loss of intricate semantics in dexterous hand motions. In contrast, our framework carefully measures the hand motion semantics with an anatomy-based semantic matrix (ASM), and transfers these semantics to the target hand motion through a novel anatomy-based semantics reconstruction network (ASRN).

\subsection{Hand-object Interaction Synthesis}

The synthesis of hand grasping given an object has been extensively studied in robotics~\cite{el20113d,bohg2013data,sahbani2012overview}. Recently, several data-driven methods have been proposed~\cite{hasson2019learning,taheri2020grab,corona2020ganhand,zhu2021toward}. Among these methods, \citet{karunratanakul2020grasping} and \citet{DBLP:conf/rss/0002ZSFZ21} propose to represent the proximity between the hand and the object as an implicit function.

Object manipulation synthesis involves dynamic hand and object interaction, which makes it more relevant to our research. Previous researchers have tackled this issue by optimizing hand poses to meet different constraints~\cite{ye2012synthesis,liu2009dextrous,mordatch2012contact,zhao2013robust}. In a recent study, \citet{DBLP:journals/tog/ZhangYSK21} employed hand-object spatial representations to learn object manipulation using motion capture data. Subsequently, \citet{DBLP:conf/eccv/ZhouBLP22} devised a different object-centric spatiotemporal representation.

However, these representations cannot capture the semantics of hand motion as they neglect the interaction between the palm and the fingers. Furthermore, these representations are explicitly designed for a given template hand mesh, which restricts their applicability to different hand models. In contrast, our ASM quantifies hand motion semantics without depending on a template mesh, allowing its application to diverse hand models.

\begin{figure*}[htbp]
  \centering
  \includegraphics[width=0.9\linewidth]{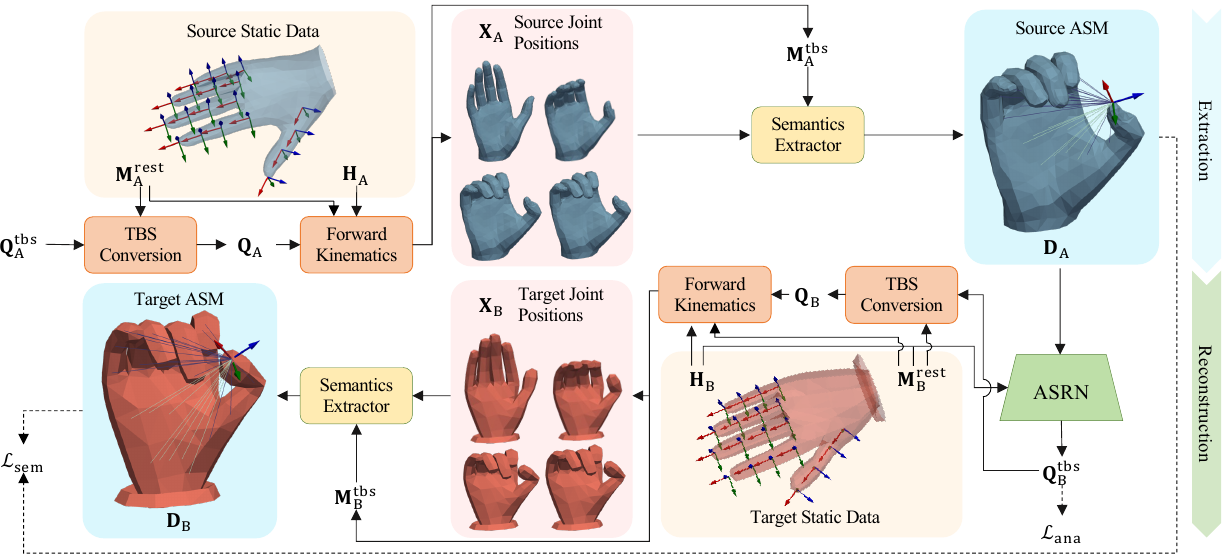}
  \caption{The figure presents an overview of the proposed pipeline consisting of two stages. The extraction stage involves the retrieval of ASM from the source hand motion. The reconstruction stage utilizes the source ASM, target hand shape parameter, and target hand anatomical parameter to reconstruct the target hand motion.}
  \label{fig:overview}
\end{figure*}

\section{Problem Formulation}
\label{sec:problem_formulation}

This paper aims to learn a mapping function $f$ that transfers the source hand motion to the target hand while preserving the semantics of the source hand motion. The inputs to the function are the source joint rotation sequence $\mathbf{Q}_{\mathrm{A}}$, the source hand shape parameter $\mathbf{H}_{\mathrm{A}}$, the source hand anatomical parameter $\mathbf{M}^{\mathrm{rest}}_{\mathrm{A}}$, the target hand shape parameter $\mathbf{H}_{\mathrm{B}}$, and the target hand anatomical parameter $\mathbf{M}^{\mathrm{rest}}_{\mathrm{B}}$. The mapping function can be formulated as follows:
\begin{equation}
  f(\mathbf{Q}_{\mathrm{A}}, \mathbf{H}_{\mathrm{A}}, \mathbf{M}^{\mathrm{rest}}_{\mathrm{A}}, \mathbf{H}_{\mathrm{B}}, \mathbf{M}^{\mathrm{rest}}_{\mathrm{B}}) \implies \mathbf{Q}_{\mathrm{B}},
\end{equation}
where $\mathbf{Q}_{\mathrm{B}}$ is the target joint rotation sequence.

\section{Methodology}
Based on the formulation in Section~\ref{sec:problem_formulation}, we have developed a framework for retargeting hand movements, as depicted in Figure~\ref{fig:overview}. We introduce a novel anatomy-based semantic matrix (ASM) based on the finger anatomical coordinate frame. By utilizing the ASM, we train an anatomy-based semantics reconstruction network (ASRN) to predict the target joint rotation sequence using the source ASM, target hand shape parameter, and target hand anatomical parameter.

In the subsequent subsections, we briefly introduce the anatomical coordinate frame of finger movements, as outlined in Section~\ref{sec:anatomical_limitations}. Next, we elaborate on the definition of the ASM in Section~\ref{sec:semantic_matrix}. Finally, we describe the framework pipeline and training details in Section~\ref{sec:semantics_preserving_retargeting}.

\begin{figure}[htbp]
  \centering
  \includegraphics[width=0.8\linewidth]{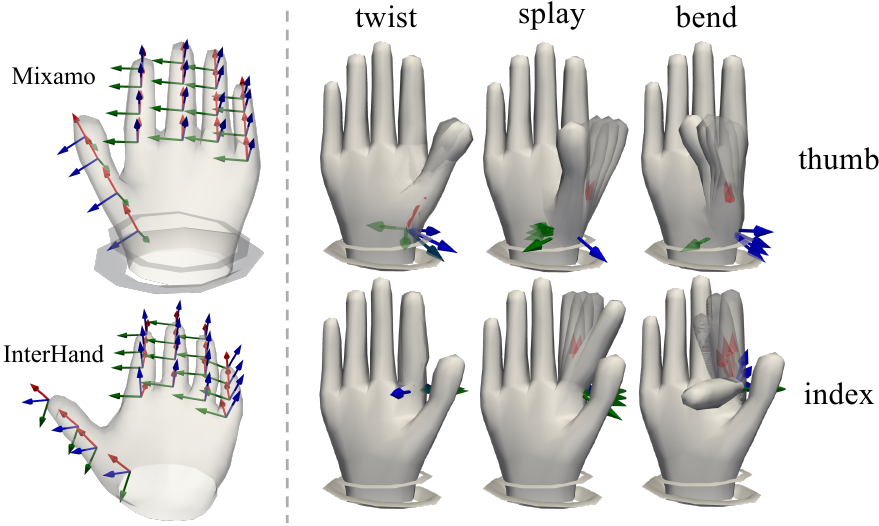}
  \caption{Left: \textit{Twist-bend-splay} frames obtained from different hand models using our annotation tool. Right: Finger movements in the \textit{twist}, \textit{splay}, and \textit{bend} directions. Note that the \textit{bend} and \textit{splay} directions of the thumb joints differ significantly from those of the other four fingers.}
  \label{fig:tbs}
  \vspace{-1em}
\end{figure}

\subsection{\textit{Twist-bend-splay} Frame}
\label{sec:anatomical_limitations}

The human hand exhibits a high degree of articulation. Directly predicting rotations of all 15 finger joints can lead to abnormal hand postures. Previous works~\cite{lin2000modeling,yang2021cpf} suggest that constraints can be applied to the finger joint rotations to prevent abnormal hand movements. \citet{yang2021cpf} extended MANO~\cite{DBLP:journals/tog/0002TB17} to develop a hand model called A-MANO incorporating anatomical constraints. A-MANO assigns a Cartesian coordinate frame, known as the \textit{Twist-bend-splay} frame, to each joint in the hand's kinematic tree. The frame's \textit{x}, \textit{y}, and \textit{z} axes align with the three revolute directions: \textit{twist}, \textit{bend}, and \textit{splay}, based on hand anatomy. Most finger joints have only one degree of freedom (DoF) along the \textit{bend} axis.

While A-MANO shows promise in estimating MANO pose during hand-object interaction, it does not apply to hand models from external sources, such as the hands of Mixamo~\cite{mixamo} characters. To mitigate this problem, we develop a tool for annotating the \textit{Twist-bend-splay} frames of different hand models. Figure~\ref{fig:tbs} demonstrates that our tool can readily provide the \textit{Twist-bend-splay} frames for hands obtained from both InterHand2.6M~\cite{Moon_2020_ECCV_InterHand2.6M} and Mixamo~\cite{mixamo}. Details of our annotation tool can be found in Appendix~\ref{sec:annotation_sup}.

\begin{figure}[htbp]
  \centering
  \includegraphics[width=0.8\linewidth]{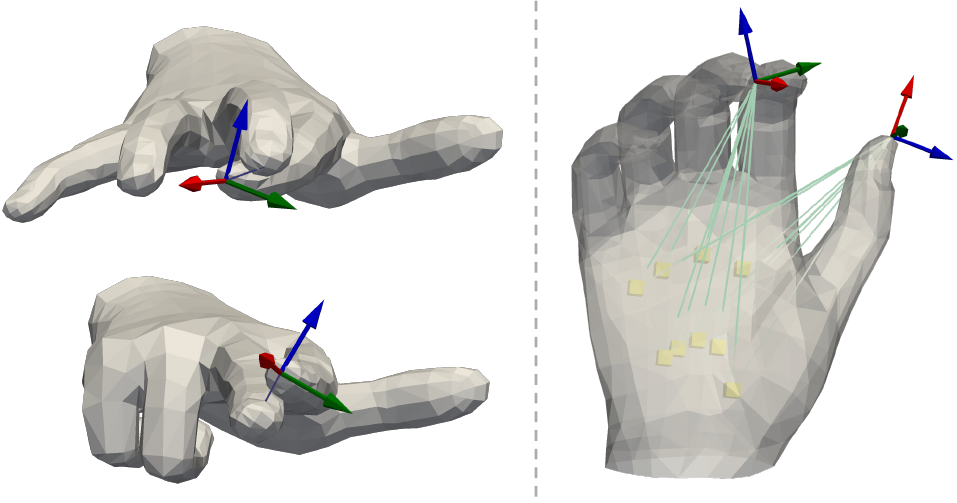}
  \caption{Left: The inter-finger semantic features capture the subtle semantics of finger movements. Right: The palm-finger semantic features capture the overall hand posture. Yellow cubes represent the palm anchors.}
  \label{fig:matrix}
  \vspace{-3em}
\end{figure}

\subsection{Anatomy-based Semantic Matrix}
\label{sec:semantic_matrix}

Our framework aims to preserve the intricate semantics while retargeting hand motions between hand models from different sources. This paper defines hand motion semantics as the spatial relationships between the fingers and the palm. Due to the absence of paired ground truth with intense semantic supervision, we introduce a novel anatomy-based semantic matrix (ASM) as a semantic measurement for hand motion retargeting. Compared to existing semantic measurements in body motion retargeting~\cite{DBLP:journals/tog/AbermanLLSCC20,zhang2023skinned,villegas2018neural} and object manipulation synthesis~\cite{DBLP:conf/eccv/ZhouBLP22,DBLP:journals/tog/ZhangYSK21}, the proposed ASM captures the intricate semantics of hand motions and can be applied to hand models from different sources without any additional cost.

Our ASM is constructed based on the \textit{twist-bend-splay} frame introduced in Section~\ref{sec:anatomical_limitations}. The crucial insight behind constructing the ASM lies in that the orientation of the \textit{twist-bend-splay} frame reveals the finger's structure. As shown in Figure~\ref{fig:matrix}, the \textit{splay} axis (blue axis) extends from the finger pulp to the back surface of the finger, while the \textit{bend} axis (green axis) stretches from the right side to the left side of the finger. The \textit{twist} axis also aligns with the finger bone. In this scenario, we can deduce the spatial relationships between the middle fingertip and the index fingertip based on the coordinates of the middle fingertip within the local \textit{twist-bend-splay} frame of the index fingertip.

The proposed ASM applies to hand models composed of five fingers, each consisting of four joints (including a dummy fingertip joint). The semantic matrix comprises two components: inter-finger semantic features and palm-finger semantic features. Formally, at time $t$, the coordinates of the $k$-th finger joint within the global frame are represented as $^\mathrm{g}\mathrm{x}_k \in \mathbb{R}^3$. $^\mathrm{g}\mathbf{M}_k$ represents the rotation matrix of the \textit{twist-bend-splay} frame of joint $k$ within the global frame. The coordinates of another joint $m$ within the local frame of joint $k$ are given by $^k\mathrm{x}_m = {^\mathrm{g}\mathbf{M}_k^{\mathrm{T}}}(^\mathrm{g}\mathrm{x}_m - {^\mathrm{g}\mathrm{x}_k})$. We define $^k\mathrm{x}_m$ as the inter-finger semantic feature of joint $m$ concerning joint $k$. Additionally, we introduce the palm-finger semantic feature to capture the overall hand posture, as depicted in Figure~\ref{fig:matrix}. Inspired by \citet{yang2021cpf}, we define nine palm anchors along the line connecting the metacarpophalangeal and wrist joints. We denote the palm-finger semantic feature of the $n$-th anchor with respect to joint $k$ as $^k\mathrm{x}_{\mathrm{p}_n} = {^\mathrm{g}\mathbf{M}_k^{\mathrm{T}}({^\mathrm{g}\mathrm{x}_{\mathrm{p}_n}} - {^\mathrm{g}\mathrm{x}_k})}$, where ${^\mathrm{g}\mathrm{x}_{\mathrm{p}_n}}$ represents the coordinates of the $n$-th anchor within the global frame. By combining the inter-finger semantic features and the palm-finger semantic features, we can construct the semantic matrix for joint $k$ as:
\begin{equation}
  \label{eq:semantics_matrix}
  ^k\mathbf{D} = [{^k\mathrm{x}_1}, {^k\mathrm{x}_2}, \dots, {^k\mathrm{x}_{20}}, {^k\mathrm{x}_{\mathrm{p}_1}}, {^k\mathrm{x}_{\mathrm{p}_2}}, \dots, {^k\mathrm{x}_{\mathrm{p}_9}}] \in \mathbb{R}^{29 \times 3}.
\end{equation}
By having semantic matrices for all 20 finger joints, we obtain the semantic measurement of the entire hand model without relying on any standard mesh template.

\subsection{Semantics-Preserving Retargeting}
\label{sec:semantics_preserving_retargeting}

The hand retargeting pipeline comprises two stages: semantic feature extraction and semantics-preserving reconstruction. We extract semantic matrices from the source hand motion during the first stage. In the second stage, we employ the anatomy-based semantics reconstruction network (ASRN) to reconstruct hand motion on the target hand model from the source ASM while preserving the source semantics. The overall pipeline is depicted in Figure~\ref{fig:overview}.

In the semantic feature extraction stage, the $T$-frame hand motion sequence in the \textit{twist-bend-splay} frame, represented as quaternions of the 15 finger joints, is denoted as $\mathbf{Q}_{\mathrm{A}}^{\mathrm{tbs}} \in \mathbb{R}^{T \times 15 \times 4}$. After converting $\mathbf{Q}_{\mathrm{A}}^{\mathrm{tbs}}$ to the global frame using the rest orientation of the joint \textit{twist-bend-splay} frames $\mathbf{M}_\mathrm{A}^\mathrm{rest} \in \mathbb{R}^{15 \times 3 \times 3}$, we obtain $\mathbf{Q}_{\mathrm{A}} \in \mathbb{R}^{T \times 15 \times 4}$. We then perform forward kinematics (FK) to derive the global coordinates of the finger joints $\mathbf{X}_{\mathrm{A}} \in \mathbb{R}^{T \times 20 \times 3}$ and the global orientation of the \textit{twist-bend-splay} frames $\mathbf{M}_\mathrm{A}^\mathrm{tbs} \in \mathbb{R}^{T \times 20 \times 3 \times 3}$. It is important to note that the FK results include the dummy fingertip joints. Additionally, the shape parameter $\mathbf{H}_\mathrm{A} \in \mathbb{R}^{h_\mathrm{A}}$ takes different forms depending on the model type. In the case of MANO models, $\mathbf{H}_\mathrm{A}$ represents shape PCA coefficients published by \citet{DBLP:journals/tog/0002TB17}, while for Mixamo models, $\mathbf{H}_\mathrm{A}$ corresponds to the normalized finger joint offsets. Finally, we extract the semantic matrices $\mathbf{D}_{\mathrm{A}} = [^1\mathbf{D}_\mathrm{A}, {^2\mathbf{D}_\mathrm{A}}, \dots, {^{20}\mathbf{D}_\mathrm{A}}] \in \mathbb{R}^{20 \times T \times 29 \times 3}$ from $\mathbf{X}_{\mathrm{A}}$ and $\mathbf{M}_\mathrm{A}^\mathrm{tbs}$ using Equation~\ref{eq:semantics_matrix}, where $^k\mathbf{D}_\mathrm{A}$ is the concatenation of $^k\mathbf{D}$ in $T$ frames.

Having obtained the semantic matrices $\mathbf{D}_{\mathrm{A}}$ from the source hand motion, we utilize our ASRN to reconstruct the target hand motion $\mathbf{Q}_{\mathrm{B}}^{\mathrm{tbs}} \in \mathbb{R}^{T \times 15 \times 4}$ on the target hand model B. A ResNet-like~\cite{he2016deep} architecture is employed. Consecutive 1D ResNet layers process the source ASM $\mathbf{D}_\mathrm{A}$. Additionally, ASRN receives the target hand shape parameter $\mathbf{H}_\mathrm{B}$ and the target hand local frame rest orientation $\mathbf{M}_\mathrm{B}^\mathrm{rest}$ as inputs. An MLP encodes $\mathbf{H}_\mathrm{B}$ and $\mathbf{M}_\mathrm{B}^\mathrm{rest}$ initially, followed by concatenation with the input of each ResNet layer. The output of the final ResNet layer is used as input for a fully-connected layer, which predicts the target hand joint rotation $\mathbf{Q}_\mathrm{B}^\mathrm{tbs}$ in target hand \textit{twist-bend-splay} frames. Next, we extract semantic matrices $\mathbf{D}_\mathrm{B}$ from the generated hand motion. In this work, hand motion semantics preservation is modeled as preserving spatial relationships between the fingers and the palm. This design defines the semantic loss $\mathcal{L}_{\mathrm{sem}}$ as the weighted cosine similarity between the source and target semantic matrices:
\begin{equation}
  \mathcal{L}_{\mathrm{sem}} = \frac{1}{T} \sum_{t=1}^T \sum_{j=1}^{20} \sum_{k=1}^{29} \omega_{jk} \frac{{\mathbf{D}_\mathrm{A}^{j,t,k}} \cdot {\mathbf{D}_\mathrm{B}^{j,t,k}}}{||{\mathbf{D}_\mathrm{A}^{j,t,k}}||_2 ||{\mathbf{D}_\mathrm{B}^{j,t,k}}||_2},
\end{equation}
where the weight $\omega_{jk}$ is defined as:
\begin{equation}
  \label{eq:weight}
  \omega_{jk} = 
  \begin{cases}
    1 + \frac{\exp(-||\mathbf{D}_\mathrm{A}^{j,t,k}||_2)}{\sum_{m=1}^{20} \exp(-||\mathbf{D}_\mathrm{A}^{j,t,m}||_2)} & \text{if $k \in \{1, 2, \dots, 20\}$} \\
    1 & \text{if $k \in \{21, 22, \dots, 29\}$}.
  \end{cases}
\end{equation}
This weighting scheme encourages the network to focus on close-finger interactions.

To mitigate abnormal hand postures generated by our network, we propose an anatomical loss, denoted as $\mathcal{L}_\mathrm{ana}$. $\mathbf{Q}_\mathrm{B}^\mathrm{tbs}$ is decomposed into three Euler angles: $\phi_\mathrm{twist}$, $\phi_\mathrm{bend}$, and $\phi_\mathrm{splay}$, aligned with the local \textit{twist-bend-splay} frame axes. Initially, we apply a penalty to $\phi_\mathrm{twist}$ for all the joints along the hand's kinematic tree. Additionally, a penalty is imposed on $\phi_\mathrm{splay}$ if it exceeds the acceptable range. Finally, we penalize the rotation angle $\mathbf{\phi}_\mathrm{bend}$ if it exceeds $\pi/2$ or falls below $0$. The anatomical loss is defined as:
\begin{equation}
  \label{eq:anatomical_loss}
  \begin{split}
    \mathcal{L}_{\mathrm{ana}} = & \frac{1}{T} \sum_{t=1}^T ( \sum_{j \in \mathrm{all}} |\phi_\mathrm{twist}^{t,j}|^2 + \sum_{j \notin \mathrm{knuckle}} |\phi_\mathrm{splay}^{t,j}|^2 \\
    & + \sum_{j \in \mathrm{knuckle}} \mathrm{max}(|\phi_\mathrm{splay}^{t,j}| - \pi/18, 0)^2 \\
    & + \sum_{j \in \mathrm{all}} \mathrm{max}(\mathbf{\phi}_\mathrm{bend}^{t,j} - \pi/2, 0)^2 + \sum_{j \in \mathrm{all}} \mathrm{min}(\mathbf{\phi}_\mathrm{bend}^{t,j}, 0)^2 ).
  \end{split}
\end{equation}

Since our network is trained on hand motion data from different hand models, the self-reconstruction supervision signals are only available when A and B belong to the same character. Therefore, ASRN is trained by minimizing the following loss function:
\begin{equation}
  \mathcal{L}_{\mathrm{total}} = \mathbbm{1}_{\mathrm{A} = \mathrm{B}} \cdot \mathrm{MSE}(\mathbf{Q}_\mathrm{A}^\mathrm{tbs}, \mathbf{Q}_\mathrm{B}^\mathrm{tbs}) - \lambda_{\mathrm{sem}} \mathcal{L}_{\mathrm{sem}} + \lambda_{\mathrm{ana}} \mathcal{L}_{\mathrm{ana}},
\end{equation}
where $\lambda_{\mathrm{sem}}$ and $\lambda_{\mathrm{ana}}$ are hyper-parameters. The indicator function $\mathbbm{1}_{\mathrm{A} = \mathrm{B}}$ takes the value $1$ if A and B belong to the same character, and $0$ otherwise.

\section{Experiments}

\subsection{Datasets}
The evaluation of our framework encompasses both the Mixamo dataset~\cite{mixamo} and the InterHand2.6M dataset~\cite{Moon_2020_ECCV_InterHand2.6M}. The Mixamo dataset comprises animations performed by various virtual characters with different shapes; however, the dataset does not guarantee consistent hand motion quality and diversity. The InterHand2.6M dataset is a comprehensive collection of hand motion data captured using a multi-view camera system and supplemented with MANO~\cite{DBLP:journals/tog/0002TB17} hand pose annotations. While the InterHand2.6M dataset offers high-quality hand motion data with considerable diversity, it has limitations regarding hand shape variations. During the training phase, we gathered 40,903 frames of hand motion data from nine distinct characters. In the testing phase, we obtained 14,316 frames of hand motion data from four different characters, ensuring that none of the testing characters were present during the network's training.

\subsection{Implementation Details}
The hyper-parameters $\lambda_{\mathrm{sem}}$ and $\lambda_{\mathrm{ana}}$ are set to $1.0$ and $0.1$ respectively. The network is trained for 100 epochs with a batch size of 64. We use the Adam optimizer~\cite{DBLP:journals/corr/KingmaB14} with the learning rate set to $10^{-4}$ to train the network. The input to the network is a sequence of 8 frames with a frame rate of 5 fps. The network is implemented in PyTorch~\cite{paszke2019pytorch} and trained on a single NVIDIA RTX 2080 Ti GPU. Further details can be found in Appendix~\ref{sec:architecture_sup}.

\subsection{Evaluation Metrics}
For hand motions with paired ground truth (GT) on different characters, we use Mean Square Error (MSE) to measure how close the retargeted joint positions are to the paired GT. In the absence of paired GT, the following metrics are used to evaluate the quality of the retargeted hand motions:
\begin{equation}
  \begin{split}
    S_{\mathrm{palm}} = \frac{1}{20 \times 9 \times T} \sum_{t=1}^T \sum_{j=1}^{20} \sum_{k=21}^{29} \frac{{\mathbf{D}_\mathrm{A}^{j,t,k}} \cdot {\mathbf{D}_\mathrm{B}^{j,t,k}}}{||{\mathbf{D}_\mathrm{A}^{j,t,k}}||_2 ||{\mathbf{D}_\mathrm{B}^{j,t,k}}||_2}, \\
    S_{\mathrm{finger}} = \frac{1}{20 \times 20 \times T} \sum_{t=1}^T \sum_{j=1}^{20} \sum_{k=1}^{20} \frac{{\mathbf{D}_\mathrm{A}^{j,t,k}} \cdot {\mathbf{D}_\mathrm{B}^{j,t,k}}}{||{\mathbf{D}_\mathrm{A}^{j,t,k}}||_2 ||{\mathbf{D}_\mathrm{B}^{j,t,k}}||_2}.
  \end{split}
\end{equation}
$S_{\mathrm{palm}}$ and $S_{\mathrm{finger}}$ represent the average cosine similarity between the retargeted hand motion and the GT hand motion. Higher values indicate better preservation of the original spatial relationships between the fingers and the palm in the retargeted hand motion.

\subsection{Qualitative Results}

\begin{figure*}[htbp]
  \centering
  \includegraphics[width=0.9\linewidth]{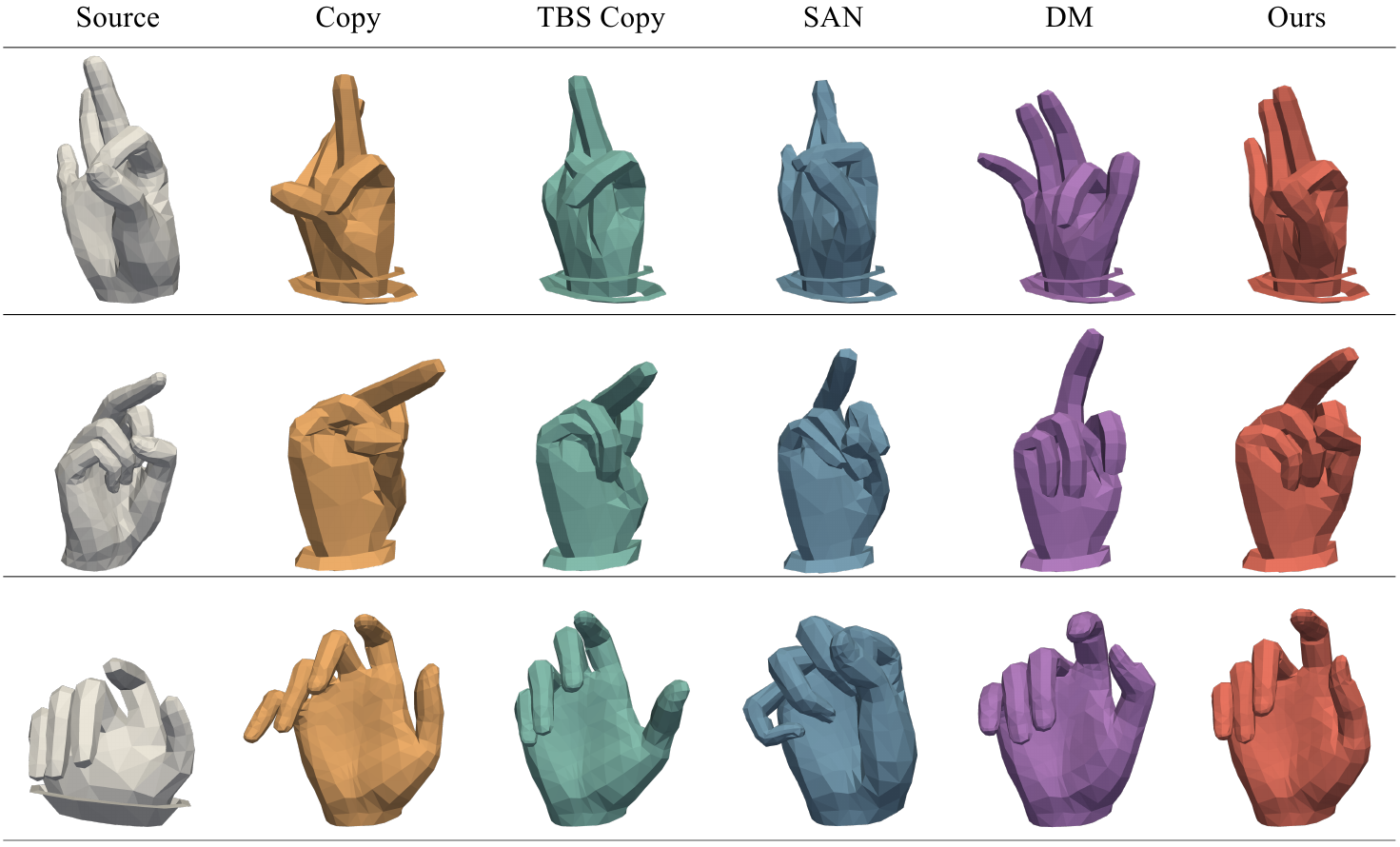}
  \caption{Qualitative comparison between the proposed framework and the state-of-the-art methods.}
  \label{fig:comparison}
\end{figure*}

The results of hand motion retargeting among hands with various shapes are depicted in Figure~\ref{fig:comparison}. The TBS Copy method copies $\mathbf{Q}_\mathrm{A}^\mathrm{tbs}$ to $\mathbf{Q}_\mathrm{B}^\mathrm{tbs}$, while the Copy method copies $\mathbf{Q}_\mathrm{A}$ to $\mathbf{Q}_\mathrm{B}$. The DM method replaces our proposed ASM with the distance matrices proposed by \citet{zhang2023skinned}. During training, the network did not encounter any of the source or target hands in the last row. Existing methods barely account for the intricate spatial relationships between the fingers and the palm, leading to inconsistent and unnatural hand motions. In contrast, our method effectively preserves the spatial relationships between the fingers and the palm, resulting in hand motions that are more natural and preserve semantics. Figure~\ref{fig:detailed_comparison} shows the detailed spatial relationships in the results of our method.

\subsection{Quantitative Results}

Table~\ref{tab:comparison} shows comparison between our method and existing body motion retargeting techniques. We compare the methods across three tasks with different sources and targets: Mixamo to Mixamo (\textbf{MX2MX}), InterHand to Mixamo (\textbf{IH2MX}), and Mixamo to InterHand (\textbf{MX2IH}). Because the Mixamo dataset provides paired GT, we use MSE to assess the quality of the retargeted hand motions for the \textbf{MX2MX} task. For the other two cross-domain tasks, we utilize $S_{\mathrm{palm}}$ and $S_{\mathrm{finger}}$ as metrics for the quality of the retargeted hand motions.

\begin{table}[htbp]
  \centering
  \caption{Comparison with the state-of-the-arts. Ours$_{w/o \mathcal{L}_{ana}}$ is the model without anatomical loss in Equation~\ref{eq:anatomical_loss}. Ours$_{w/o \text{weight}}$ is the model without the weight scheme in Equation~\ref{eq:weight}.}
  \label{tab:comparison}
  \begin{tabular}{c|c|cc|cc}
      \toprule
       \multirow{2}*{\textbf{Methods}} & \textbf{MX2MX} & \multicolumn{2}{|c|}{\textbf{IH2MX}} & \multicolumn{2}{|c}{\textbf{MX2IH}} \\ \cline{2-6}
       & $\mathrm{MSE}{\downarrow}$ & $S_{\mathrm{palm}}{\uparrow}$ & $S_{\mathrm{finger}}{\uparrow}$ & $S_{\mathrm{palm}}{\uparrow}$ & $S_{\mathrm{finger}}{\uparrow}$ \\
       \midrule
       Copy & \textbf{4.76e-12}  & 0.923 & 0.851 & 0.941 & 0.872 \\
       TBS Copy & 0.155 & 0.960 & 0.883 & 0.968 & 0.891 \\
       \midrule
       SAN~\cite{DBLP:journals/tog/AbermanLLSCC20} & 3.134 & 0.866 & 0.820 & 0.034 & 0.475 \\
       DM~\cite{zhang2023skinned} & 2.788 & 0.888 & 0.832 & 0.891 & 0.878 \\
       \midrule
       Ours$_{w/o \mathcal{L}_{ana}}$ & 0.276 & \textbf{0.983} & \textbf{0.932} & \textbf{0.985} & \textbf{0.935} \\
       Ours$_{w/o \text{weight}}$ & 0.420 & 0.972 & 0.922 & 0.980 & 0.927 \\
       Ours & 0.452 & 0.971 & 0.925 & 0.978 & 0.929 \\
       \bottomrule
  \end{tabular}
\end{table}

Because the Mixamo dataset may create a new character with an archived motion by using motion copy, the Copy method has the lowest MSE. However, as the qualitative results reveal, this does not mean the motion copy is optimal. Our method achieves a reduction in MSE of 85.6\% and 83.8\% compared to SAN~\cite{DBLP:journals/tog/AbermanLLSCC20} and DM~\cite{zhang2023skinned}, which utilize distinct semantic measurements. Additionally, our method achieves the highest $S_{\mathrm{palm}}$ and $S_{\mathrm{finger}}$ in the \textbf{IH2MX} and \textbf{MX2IH} tasks, indicating that its superior ability to preserve the original spatial relationships between the fingers and the palm during the retargeted hand motion. This observation suggests that our proposed ASM outperforms the distance matrices~\cite{zhang2023skinned} and the implicit measurement learned in SAN~\cite{DBLP:journals/tog/AbermanLLSCC20}.

\subsection{User Study}

We conduct a user study to evaluate the performance of our framework against Copy, SAN~\cite{DBLP:journals/tog/AbermanLLSCC20}, and DM~\cite{zhang2023skinned}. We invited 26 participants and showed them six static hand posture pictures and six hand motion videos. Each picture and video contains one source motion and four anonymous results. Participants were instructed to rank the pictures and videos based on three aspects: preservation of static posture semantics (PS), preservation of motion semantics (MS), and motion quality (MQ), from best to worst. The average rankings are presented in Table~\ref{tab:user_study}. Overall, our method achieved the best performance in all three aspects.

\begin{table}[htbp]
  \centering
  \caption{Ranking results of the user study. We invite 26 participants to compare the retargeting results from three aspects: static posture semantics (PS), motion semantics (MS), and motion quality (MQ).}
  \label{tab:user_study}
  \begin{tabular}{c|c|c|c}
    \toprule
    \multirow{2}*{\textbf{Methods}} & \multicolumn{3}{|c}{\textbf{Ranking}} \\ \cline{2-4}
    & PS $\downarrow$ & MS $\downarrow$ & MQ $\downarrow$ \\
    \midrule
    Copy & 3.17 & 3.31 & 3.33 \\
    SAN~\cite{DBLP:journals/tog/AbermanLLSCC20} & 2.95 & 2.99 & 3.05 \\
    DM~\cite{zhang2023skinned} & 2.70 & 2.53 & 2.46 \\
    Ours & \textbf{1.17} & \textbf{1.17} & \textbf{1.16} \\
    \bottomrule
  \end{tabular}
\end{table}

\section{Conclusion}
In this paper, we propose the problem of semantics-preserving hand motion retargeting. We encode the spatial relationships between the fingers and the palm using anatomy-based semantic matrices (ASM). We train an anatomy-based semantics reconstruction network (ASRN) to retarget the motion semantics of the source hand onto the target hand, utilizing the source ASM. We evaluate our framework on both intra-domain and cross-domain retargeting tasks. Our method demonstrates superior performance to existing motion retargeting methods, both qualitatively and quantitatively.

\section*{Acknowledgements}
This work is supported by the National Key R\&D Program of China under Grant No. 2021QY1500, the State Key Program of the National Natural Science Foundation of China (NSFC) (No.61831022). It is also supported in part by the NSFC under Grant No. 62222606 and 62076238. Thank Cuiwen for her support and encouragement.

\bibliographystyle{ACM-Reference-Format}
\bibliography{main}

\newpage
\appendix

\section{\textit{Twist-bend-splay} Frame Annotation}
\label{sec:annotation_sup}

\begin{figure}[htbp]
  \centering
  \includegraphics[width=\linewidth]{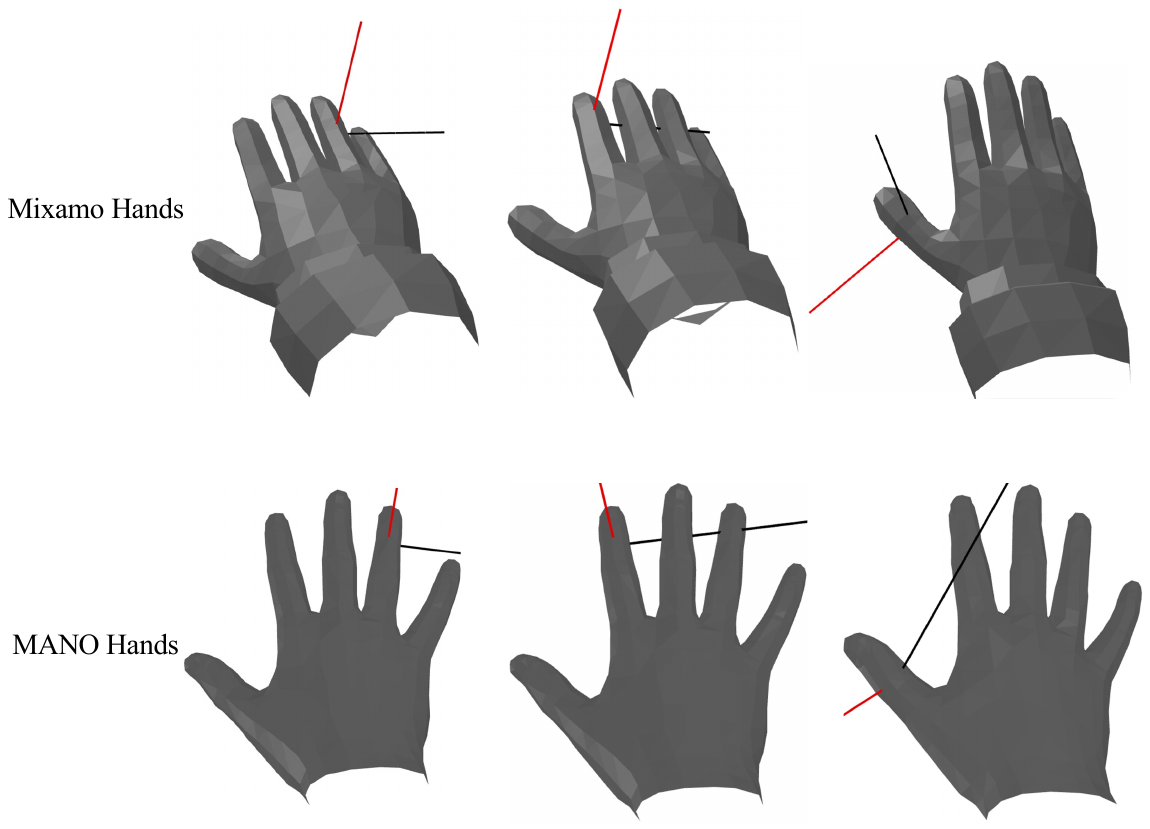}
  \caption{Our annotation tool allows the user to adjust the \textit{splay} axis (red axis) and \textit{bend} axis (black axis) directions for Mixamo and MANO hands.}
  \label{fig:annotation}
\end{figure}

This section presents our frame annotation tool for \textit{Twist-bend-splay}. A previous study by \citet{yang2021cpf} introduced A-MANO, a hand model that incorporates \textit{Twist-bend-splay} frames. A-MANO, an extension of MANO, is limited in its applicability to other hand models. This paper presents the implementation of a versatile frame annotation tool for \textit{Twist-bend-splay}, applicable to any hand model with five fingers and 15-finger joints. The annotation tool can semi-automatically derive the frame orientation of finger joints for \textit{Twist-bend-splay} from the model's kinematic tree and mesh information.

Specifically, our annotation tool first computes the twist axis $\mathbf{n}_\mathrm{twist}$ as the vector from the child of the current joint to the joint itself. Next, we project rays onto the normal plane defined by $\mathbf{n}_\mathrm{twist}$ and perform ray-mesh queries. The ray-mesh hit locations on the mutually perpendicular axes $\mathbf{n}_\mathrm{splay}$ and $\mathbf{n}_\mathrm{bend}$ are denoted as $\mathbf{p}_\mathrm{splay}$ and $\mathbf{p}_\mathrm{bend}$, respectively. $\mathbf{m}_\mathrm{splay}$ and $\mathbf{m}_\mathrm{bend}$ represent the normal vectors of the mesh at $\mathbf{p}_\mathrm{splay}$ and $\mathbf{p}_\mathrm{bend}$, respectively. We iterate through all the possible axis directions and minimize the following loss function:
\begin{equation}
  \mathcal{L}_\mathrm{annotate} = -\mathbf{n}_\mathrm{spaly} \cdot \mathbf{m}_\mathrm{splay} - \mathbf{n}_\mathrm{bend} \cdot \mathbf{m}_\mathrm{bend} + \frac{||\mathbf{p}_\mathrm{splay} - \mathbf{o}||_2}{||\mathbf{p}_\mathrm{bend} - \mathbf{o}||_2},
\end{equation}
where $\mathbf{o}$ is the location of the corresponding finger joint. The underlying insight of $\mathcal{L}_\mathrm{annotate}$ is that the fingers are narrower from top to bottom but wider from left to right. Therefore, we minimize $\frac{||\mathbf{p}_\mathrm{splay} - \mathbf{o}||_2}{||\mathbf{p}_\mathrm{bend} - \mathbf{o}||_2}$. Moreover, we aim to align the axes with the mesh normals, thus maximizing $\mathbf{n}_\mathrm{spaly} \cdot \mathbf{m}_\mathrm{splay} + \mathbf{n}_\mathrm{bend} \cdot \mathbf{m}_\mathrm{bend}$. Finally, our annotation tool displays the frames of \textit{Twist-bend-splay} on the hand model, as depicted in Figure~\ref{fig:annotation}. If needed, the user can manually adjust the orientation of the \textit{splay} and \textit{bend} axes.

\section{Network Architecture and Training Details}
\label{sec:architecture_sup}

\begin{figure}[htbp]
  \centering
  \includegraphics[width=0.4\linewidth]{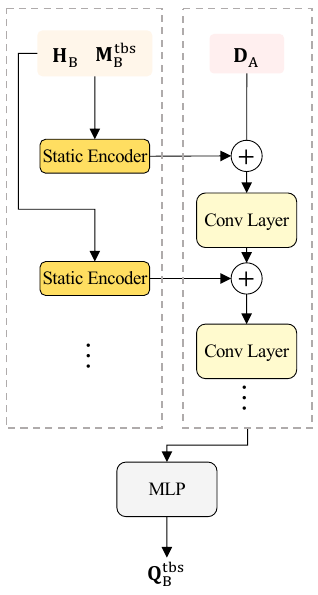}
  \caption{The network architecture of the proposed ASRN.}
  \label{fig:architecture}
\end{figure}

As depicted in Figure~\ref{fig:architecture}, the proposed Action Sequence Reconstruction Network (ASRN) architecture comprises two main components: the static encoders and the motion reconstruction convolutional network. Each static encoder consists of one MLP layer and two ResNet-like convolutional layers. The motion reconstruction convolutional network is composed of four ResNet-like convolutional layers. The input to each layer concatenates the output from the previous layer and the output from the corresponding static encoder. The ASRN takes the source ASM denoted as $\mathbf{D}_\mathrm{A}$ as input and generates the target joint rotation denoted as $\mathbf{Q}_\mathrm{B}^\mathrm{tbs}$ as output. To train the ASRN, we employ the Adam optimizer~\cite{DBLP:journals/corr/KingmaB14} with a learning rate of $10^{-4}$ and a batch size of 64. The ASRN is trained for 100 epochs.

Since the shape parameter $\mathbf{H}_\mathrm{B} \in \mathbb{R}^{h_\mathrm{B}}$ varies based on the model type, we train an ASRN for each specific form of the shape parameter. In this study, we introduce two ASRNs specifically for MANO and Mixamo. For MANO, $\mathbf{H}_\mathrm{B}$ is represented as a 10-dimensional vector, while Mixamo represents a 45-dimensional vector. The ASRNs for both MANO and Mixamo are trained using identical hyperparameters. Each network is trained on the InterHand2.6M and Mixamo datasets, but with distinct target hand models.

\begin{figure}[ht]
  \centering
  \includegraphics[width=\linewidth]{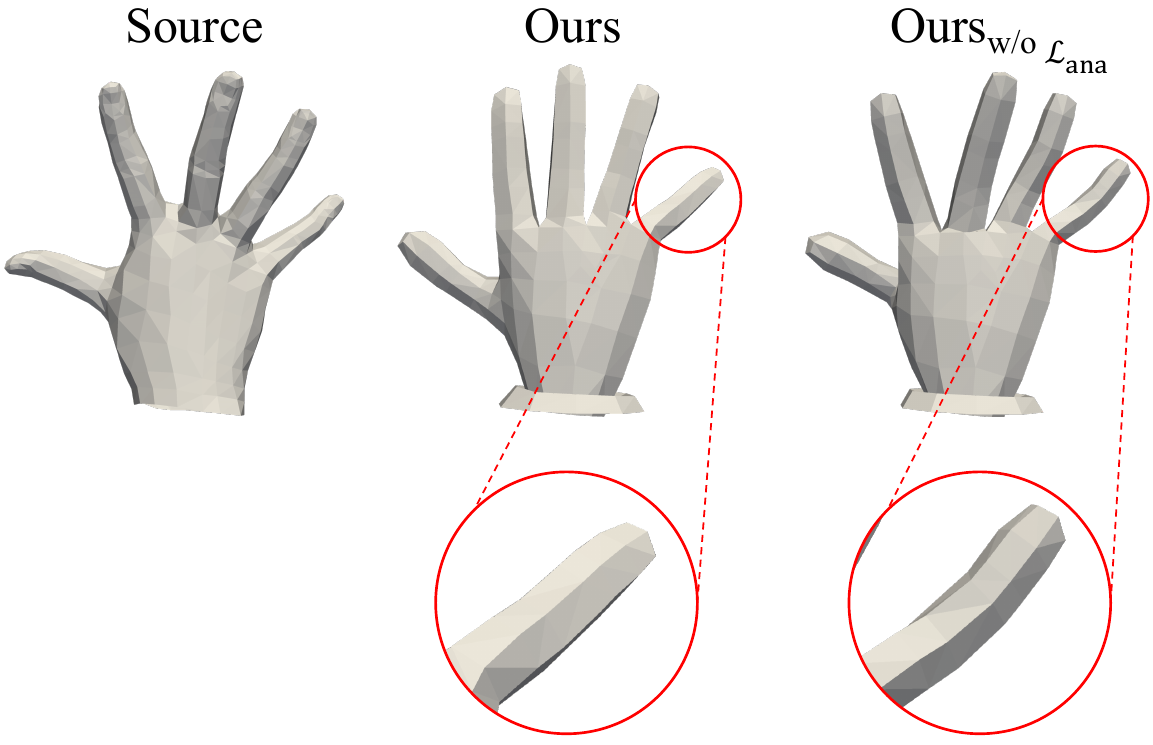}
  \caption{Comparison of results with and without the inclusion of the anatomical loss $\mathcal{L}_\mathrm{ana}$.}
  \label{fig:abltaion_ana}
\end{figure}

\begin{figure}[ht]
  \centering
  \includegraphics[width=\linewidth]{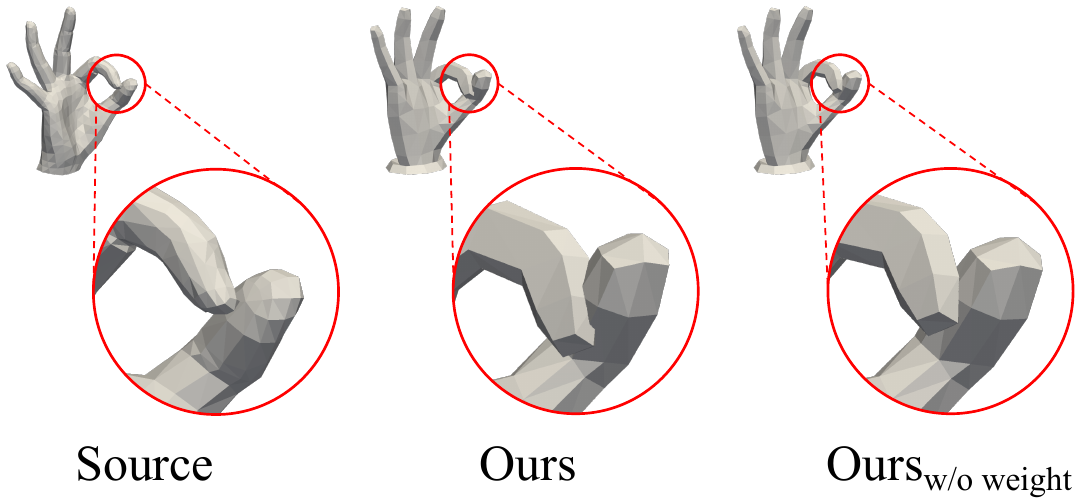}
  \caption{Comparison of results with and without the weighting scheme described in Equation~\ref{eq:weight}.}
  \label{fig:abltaion_weight}
\end{figure}

\section{Ablation Study}
The qualitative results of two ablated versions of our methods are illustrated in Figure~\ref{fig:abltaion_ana} and Figure~\ref{fig:abltaion_weight}.

Figure~\ref{fig:abltaion_ana} compares the results with and without the inclusion of the anatomical loss $\mathcal{L}_\mathrm{ana}$. Excluding $\mathcal{L}_\mathrm{ana}$ leads to a higher occurrence of unnatural finger poses, such as the abnormal splay of the interphalangeal joint of the little finger.

Figure~\ref{fig:abltaion_weight} compares the results with and without using the weighting scheme described in Equation~\ref{eq:weight}. The weighting scheme promotes the network's attention toward proximal joint interactions. Consequently, the whole model produces a motion where the thumb pulp contacts the index fingertip, while the ablated model fails to achieve this contact.

\section{Supplementary Qualitative Results}
\label{sec:comparison_sup}

\begin{figure}[htbp]
  \centering
  \includegraphics[width=\linewidth]{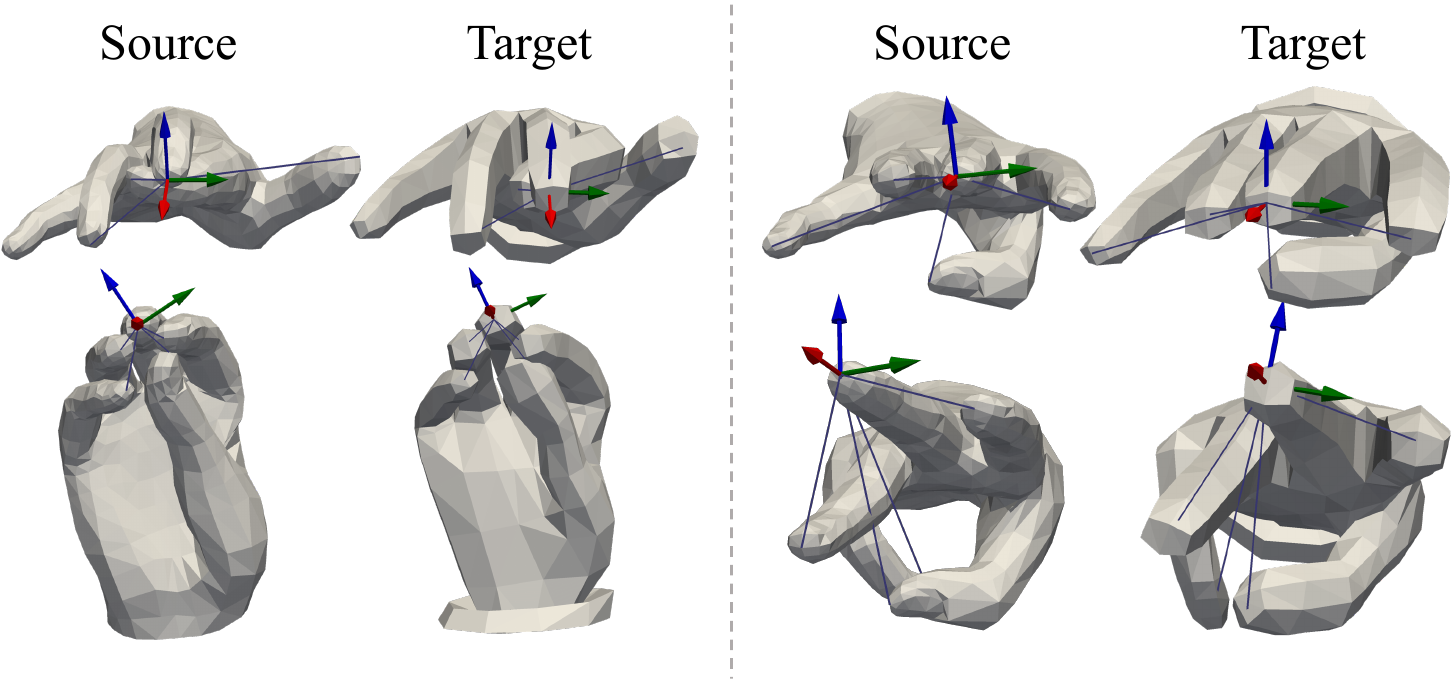}
  \caption{Our framework maintains precise spatial relationships among the fingers.}
  \label{fig:detailed_comparison}
\end{figure}

Figure~\ref{fig:detailed_comparison} presents additional qualitative results of our method. Our approach effectively preserves accurate hand motion semantics.

\end{document}